\documentclass[conference]{IEEEtran}
\IEEEoverridecommandlockouts
\usepackage{cite}
\usepackage{amsmath,amssymb,amsfonts}
\usepackage{algorithmic}
\usepackage{graphicx}
\usepackage{textcomp}
\usepackage{xcolor}
\usepackage{booktabs}
\usepackage{multirow}
\usepackage{array} 
\usepackage[most]{tcolorbox} 
\usepackage[colorlinks=true, linkcolor=black, citecolor=black, urlcolor=black]{hyperref}

\definecolor{darkblue}{rgb}{0,0,0.8}

\tcbuselibrary{listings, breakable}

\def\BibTeX{{\rm B\kern-.05em{\sc i\kern-.025em b}\kern-.08em
    T\kern-.1667em\lower.7ex\hbox{E}\kern-.125emX}}
\begin{document}

\title{\vspace{-1.3em}HARP-LoRA: Hierarchy-Conditioned Rank Projection for Low-Rank Adaptation in Scientific Table Understanding }


\author{
Hossein Sholehrasa\textsuperscript{1,2},
Mohammadhossein Amini\textsuperscript{3},
Doina Caragea\textsuperscript{2},
Remya Ampadi Ramachandran\textsuperscript{1,4},\\
Dan Wu\textsuperscript{3},
Zihang Shi\textsuperscript{5},
Zelin Li\textsuperscript{6},
Lindsey Hulbert\textsuperscript{1,7},
Eder Martinez\textsuperscript{2},
Xuan Xu\textsuperscript{1,8},
Majid Jaberi-Douraki\textsuperscript{1,4,*}
\thanks{*Corresponding Author: \texttt{jaberi@k-state.edu}}
\\[1ex]\\ \small
\textsuperscript{1}1DATA Consortium and FARAD Program, Kansas State University, Olathe, KS, USA; 
\textsuperscript{2}Department of Computer Science, Kansas State \\ University, Manhattan, KS, USA;

\textsuperscript{3}Olin Business School, Washington University in St. Louis, St. Louis, Missouri, USA;
\\
\textsuperscript{4}Department of Mathematics, Kansas State University, Manhattan, KS, USA;

\textsuperscript{5}Institute for Health Informatics, University of Minnesota \\ Medical School, Minneapolis, MN, USA;
\textsuperscript{6}Miami Herbert Business School, University of Miami, Coral Gables, Florida, USA;
\\
\textsuperscript{7}Department of Animal Sciences and Industry, Kansas State University, Manhattan, KS, USA;
\\
\textsuperscript{8}Department of Statistics, Kansas State University, Manhattan, KS, USA
\vspace{-1.2em}
}




\maketitle

\begingroup
\renewcommand\thefootnote{}
\footnotetext{
\scriptsize

© 2026 IEEE. Personal use of this material is permitted. Permission from IEEE must be obtained for all other uses, in any current or future media, including reprinting/republishing this material for advertising or promotional purposes, creating new collective works, for resale or redistribution to servers or lists, or reuse of any copyrighted component of this work in other works. This is the author's accepted version of a paper accepted for publication in the Proceedings of the 2026 IEEE 38th International Conference on Tools with Artificial Intelligence (ICTAI).
}
\endgroup

\begin{abstract}
Tabular data is widely used in scientific and real-world applications, where complex table layouts, including multi-level headers, merged cells, captions, and footnotes, can make reliable information extraction difficult. This challenge is particularly important in pharmacokinetics (PK), where numerical values must be correctly associated with the corresponding drug, species, dose, administration route, unit, and experimental condition. Standard parameter-efficient fine-tuning methods such as Low-Rank Adaptation (LoRA) reduce adaptation cost, but they do not explicitly incorporate the relationships between table values and their surrounding headers into the adaptation process. To address this limitation, we introduce HARP-LoRA, a hierarchy-conditioned rank projection method for low-rank adaptation in scientific table understanding. HARP-LoRA extends projected LoRA by constructing token-specific intrinsic representations from the row- and column-header hierarchies associated with each table data token. These representations condition the low-rank down-projection, allowing the adapter to incorporate hierarchical context while keeping the pretrained model and random projection tensors frozen. We evaluate HARP-LoRA on 950 manually annotated PK tables collected from 587 scientific articles and on WikiTableQuestions (WikiTQ) using Llama~3.2 3B, Llama~3.1 8B, and Gemma~3 27B. With Llama~3.1 8B, HARP-LoRA achieves a PK F1-score of 57.64\% and a WikiTQ accuracy of 53.15\%, improving over LoRA by 1.42 and 6.56 points, respectively. With Gemma~3 27B, it achieves the best PK F1-score of 70.63\% and WikiTQ accuracy of 56.87\%. These results are obtained with only 172K, 197K, and 547K trainable parameters, substantially fewer than LoRA and TableLoRA. Overall, the results indicate that conditioning lightweight adapters on table-header hierarchies can improve both information extraction from hierarchy-rich scientific tables and reasoning over flatter general-domain tables while maintaining strong parameter efficiency. Code is available at: \href{https://github.com/hosseinsholehrasa/HARP-LoRA}{\textcolor{darkblue}{https://github.com/hosseinsholehrasa/HARP-LoRA}}
\end{abstract}


\section{Introduction}
Tabular data is widely used across many real-world applications, making table understanding and reasoning an important research problem. Benchmarks such as WikiTableQuestions (WikiTQ) \cite{pasupat2015compositional}, HiTab \cite{cheng2022hitab}, and TabFact \cite{chen2019tabfact} evaluate models on tasks such as retrieval, aggregation, and verification over structured tables. One domain where accurate table understanding is especially critical is pharmacokinetics (PK), the study of how drugs are absorbed, distributed, metabolized, and eliminated by the body \cite{maunsell2021farad}. PK parameters such as half-life, clearance, and area under the curve are important for understanding drug behavior and supporting dosing, safety, and regulatory decisions \cite{riviere2018veterinary,sholehrasa2025predictive}. They are also important for resources such as Food Animal Residue Avoidance Databank (FARAD) \cite{farad}, which uses accurate PK information to support drug-use and withdrawal-interval recommendations for food-producing animals. In practice, these values are often reported in complex scientific tables with multi-level headers, merged cells, and inconsistent terminology, making reliable automated extraction particularly challenging \cite{sholehrasa2025autopk, golmohammadi2025comprehensive}. Across both general table benchmarks and scientific PK tables, accurate understanding depends on correctly interpreting relationships among cells, rows, columns, and headers.

To address this structural challenge, earlier work such as TAPAS \cite{herzig2020tapas} and TURL \cite{deng2022turl} explicitly modeled table structure through row and column information, specialized representations, and dedicated pretraining. While these methods improve table understanding by explicitly modeling structural relationships, they often rely on table-specific architectures and representations together with dedicated pretraining or task-specific data \cite{yin2020tabert, wang2021tuta}. In contrast, large language models (LLMs) provide a more general foundation, with strong language understanding and reasoning capabilities that can be adapted across diverse table tasks without requiring a separate architecture for each task \cite{zhang2024tablellama}. Building on this flexibility, recent LLM-based approaches adapt general-purpose models to tabular data through table-specific fine-tuning, serialized table representations, and task-oriented processing strategies \cite{li2024tablegpt, zhang2025tablellm, sholehrasa2025autopk}. However, extensive task-specific fine-tuning becomes increasingly expensive as model size grows, motivating more efficient ways to adapt LLMs to structured table tasks.

Parameter-efficient fine-tuning (PEFT) addresses this need by adapting pretrained models through a small number of trainable parameters while keeping the base model frozen \cite{peft}. Low-Rank Adaptation (LoRA) \cite{hu2022lora} keeps pretrained weights frozen and learns task-specific updates through two small trainable low-rank matrices, greatly reducing the number of parameters optimized during fine-tuning. Despite their efficiency, PEFT methods such as LoRA do not explicitly account for the structure of tabular inputs. When a table is processed by an LLM, its two-dimensional structure is typically serialized into a one-dimensional token sequence \cite{sui2024tablemeetsllm}. LoRA adapts the resulting token representations without explicitly conditioning on their row, column, or header relationships, which can limit how well structural context is preserved in complex tables with multi-level headers. TableLoRA \cite{he2025tablelora} takes an important step toward structure-aware PEFT by assigning table tokens row and column information during adaptation, allowing the low-rank update to reflect their two-dimensional positions. However, these coordinates describe where a token is located, not the semantic relationships formed by multi-level row and column headers. Designing parameter-efficient adapters that incorporate such hierarchical structure directly therefore remains underexplored.

These limitations motivate a structure-aware PEFT method that incorporates hierarchical table semantics directly into adaptation. We propose HARP-LoRA, a \textbf{H}ier\textbf{A}rchy-conditioned \textbf{R}ank \textbf{P}rojection method for low-rank adaptation in scientific table understanding. HARP-LoRA derives token-specific representations from the corresponding row- and column-header hierarchies and uses them to condition the low-rank projection while keeping the trainable parameter budget small. We evaluate HARP-LoRA on a hierarchy-rich PK information-extraction dataset and on WikiTQ, which primarily contains flatter table structures. The main contributions of this work are as follows:
\begin{itemize}

\item We introduce HARP-LoRA, a hierarchy-conditioned PEFT method that generates token-specific intrinsic representations from row- and column-header hierarchies to condition low-rank adaptation.

\item We develop a parameter-efficient representation of multi-level headers using header-cell mean pooling, level-preserving adaptive pooling, and root mean square (RMS) normalization, allowing hierarchical structural information to be incorporated directly into the adapter.

\item We evaluate HARP-LoRA on 950 manually annotated PK tables and WikiTQ using three language-model backbones, comparing against LoRA, TableLoRA, and projected LoRA.

\item We show that HARP-LoRA achieves comparable or improved performance across the evaluated settings, outperforming existing baselines in several cases while using substantially fewer trainable parameters than LoRA and TableLoRA.

\end{itemize}

\section{Related Work}

\subsection{Table Structure Modeling and LLM Reasoning}

Prior work has explicitly incorporated table structure through specialized representations and pretraining objectives. TaBERT \cite{yin2020tabert} jointly encodes textual context and table content, TAPAS \cite{herzig2020tapas} incorporates row, column, and rank embeddings, TURL \cite{deng2022turl} uses structure-aware modeling of table relationships, and OmniTab \cite{jiang2022omnitab} pretrains on natural and synthetic question--table pairs. More recent work adapts general-purpose LLMs directly to tabular tasks. TableGPT \cite{li2024tablegpt} and TableLLM \cite{zhang2025tablellm} fine-tune LLMs across diverse table understanding and reasoning tasks, while studies such as Table Meets LLM \cite{sui2024tablemeetsllm} examine how table serialization affects LLM understanding of structured data. In domain-specific settings, AutoPK \cite{sholehrasa2025autopk} improves pharmacokinetic table extraction through structural filtering and reconstruction before LLM-based reasoning.

Together, these approaches incorporate table structure through specialized representations, pretraining, fine-tuning, serialization, or external structural processing. Earlier structure-aware methods often require changes to the underlying transformer, such as additional embeddings, modified attention, or dedicated pretraining, making them less straightforward to transfer to general-purpose LLMs. More recent LLM-based approaches reduce this dependency, but typically represent structure through serialized inputs, prompting, retrieval, or preprocessing, leaving the model to recover much of the underlying table organization from the input sequence. This motivates lightweight adaptation methods that can incorporate table structure directly during fine-tuning without modifying the base LLM architecture.

\subsection{Table-Specific PEFT}

PEFT provides such a setting by specializing pretrained LLMs through a small number of trainable parameters while keeping the base weights frozen, reducing memory, storage, and training costs for downstream adaptation \cite{peft, jaberi2025hysim}. LoRA \cite{hu2022lora} has motivated a broad range of extensions that modify different aspects of its low-rank parameterization. AdaLoRA \cite{zhang2023adalora} dynamically allocates rank, while PiSSA \cite{meng2024pissa} uses principal singular components of pretrained weights for initialization. LoRA-XS \cite{balazy2405lora} learns a small matrix between fixed low-rank factors derived from pretrained weights, while projected LoRA \cite{projected_lora} constructs LoRA factors from fixed random projection tensors and optimizes only low-dimensional intrinsic vectors.

These methods improve different aspects of PEFT, including rank allocation, initialization, and parameter compression, but their adaptation mechanisms are not designed around the structure of tabular inputs. TableLoRA \cite{he2025tablelora} is a table-specific extension that incorporates special table tokens together with row and column positional information during low-rank adaptation, allowing the adapter to explicitly account for two-dimensional table coordinates. However, coordinate information alone does not explicitly capture the semantic dependencies induced by multi-level row and column headers. This distinction is particularly important for scientific tables, where the meaning of a cell may depend on several levels of surrounding headers rather than only its row and column position.

\subsection{Scientific Table Understanding in Pharmacokinetics}

Automating PK information extraction has become increasingly important as the volume of scientific literature grows and manual curation becomes harder to scale. PK values are often reported in complex scientific tables and must be interpreted together with contextual information such as drug, species, dose, route, units, and experimental conditions. In many cases, these relationships are distributed across multi-level row and column headers, so the meaning of a value depends on its position within the surrounding header hierarchy rather than on the cell alone. Reliable extraction therefore depends on understanding both the values and the structural relationships that connect them to their context. Traditionally, PK data have been manually curated by domain experts \cite{pkdb}. WCPK reduces the effort of finding relevant studies by automatically collecting PK-related publications, but it does not extract information from the tables themselves \cite{ramachandran2023automated}.

Building on literature collection, subsequent work has moved toward identifying which tables contain relevant PK information and, more recently, extracting the information within them. Smith et al. developed a classification pipeline to identify tables containing in-vivo PK parameters and study-population characteristics \cite{smith2025automated}, providing a filtering step before downstream extraction. AutoPK \cite{sholehrasa2025autopk} moves further toward extraction from complex PK tables by identifying parameter variants, filtering relevant regions, and reconstructing standardized representations before LLM-based extraction. Despite this progress, existing approaches primarily address different stages of the extraction pipeline through retrieval, classification, preprocessing, or serialized LLM inputs, while hierarchical relationships in scientific tables remain only indirectly represented during model adaptation.

\section{Problem Definition and Motivation}

LLMs process input as a one-dimensional sequence of tokens, which is well suited to ordinary text but does not directly preserve the two-dimensional organization of tables. After serialization, table tokens retain their textual content but their row--column relationships, including links to row labels, column headers, and higher-level headers, must be inferred from the sequence. LoRA operates on the same serialized representation and therefore does not explicitly incorporate these structural relationships into its adaptation. Adding such structure directly to the LLM architecture would typically require architectural changes and additional pretraining or substantial retraining, increasing computational and financial cost.

This limitation is particularly important for scientific tables such as PK tables, where a value must remain correctly associated with its parameter, drug, species, dose, route of administration, unit, and experimental condition. Incorrect associations can lead to misleading extraction, which is especially critical for resources such as FARAD \cite{farad}, where accurate PK information supports veterinary drug-use and withdrawal-interval recommendations. Existing approaches generally encode table structure through serialization, positional information, preprocessing, or table-aware PEFT based on row and column coordinates, but do not directly model multi-level header hierarchies within the adaptation mechanism.

We therefore propose a parameter-efficient framework that conditions low-rank adaptation on the corresponding row- and column-header hierarchies of each table token. The method generates token-specific LoRA projections while keeping the pretrained model frozen. We evaluate it on a hierarchy-rich PK dataset and on WikiTQ, which primarily contains flatter table structures.

\section{Methodology}
\label{sec:method}

\begin{figure*}[!t]
    \vspace{-2em}
    \centering
    \includegraphics[width=1.\textwidth, height=0.30\textheight]{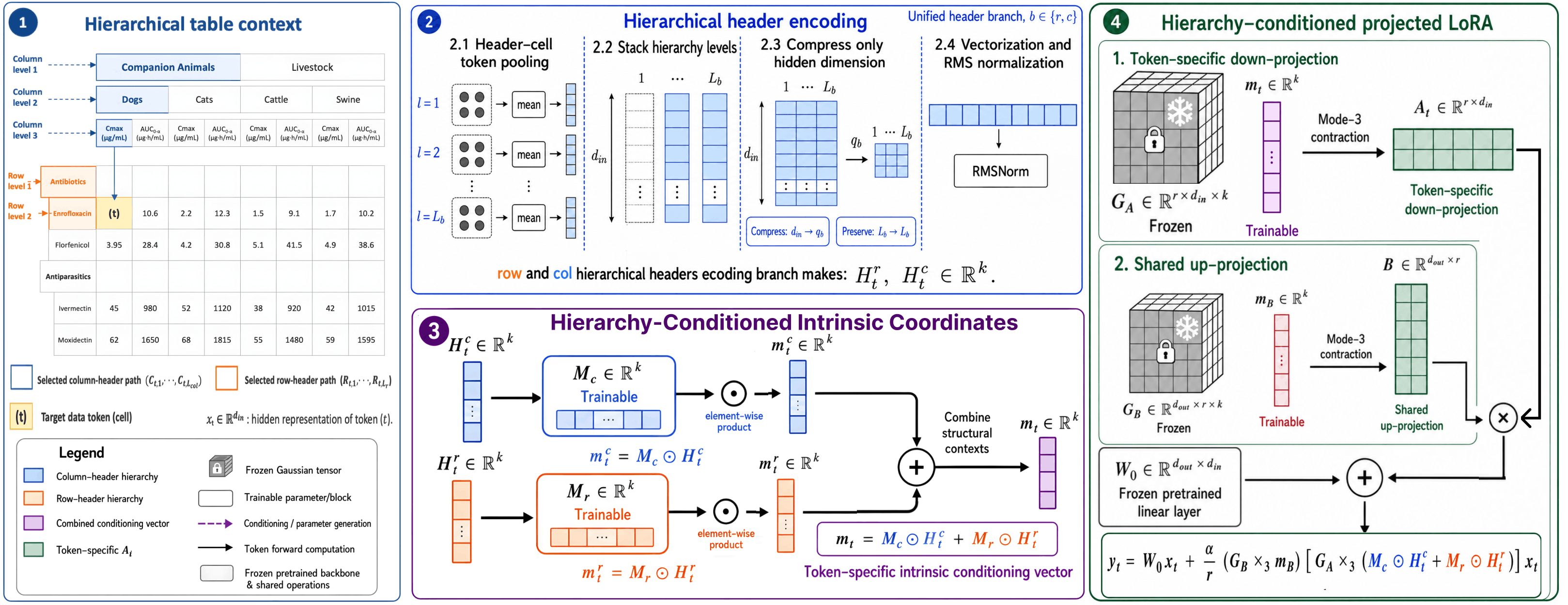}
    \vspace{-1.7em}
    \caption{Overview of HARP-LoRA. For each table data token, the corresponding hierarchical row and column headers are identified and encoded separately. Their representations are combined using separate trainable vectors to produce a token-specific intrinsic vector, which conditions the projected LoRA down-projection. The resulting low-rank representation is mapped back to the model hidden space using a shared up-projection.}
    \label{fig:table_overview_process}
    \vspace{-1em}

\end{figure*}

To incorporate table structure into the adaptation process, HARP-LoRA begins by identifying the row- and column-header hierarchies associated with each table token. For a given table token, we determine the row and column headers connected to its position and encode the two hierarchical paths separately. These representations are then combined into a conditioning vector that captures the token's structural context. The conditioning vector is used to construct a token-specific LoRA down-projection, allowing the adaptation to vary according to the token's position within the table hierarchy. The resulting low-dimensional representation is then mapped back to the model hidden space through a shared LoRA up-projection. Figure~\ref{fig:table_overview_process} illustrates the overall workflow.

\subsection{Preliminaries}
\noindent
\textbf{LoRA:}
We first introduce the LoRA formulation for a pretrained linear layer:
\begin{equation}
    \Delta \mathbf{y}_t
    =
    \frac{\alpha}{r}
    \mathbf{B}\mathbf{A}\mathbf{x}_t,
\end{equation}
where $\mathbf{x}_t \in \mathbb{R}^{d_{\mathrm{in}}}$ is the hidden
representation of token $t$,
$\mathbf{A} \in \mathbb{R}^{r \times d_{\mathrm{in}}}$ is the
down-projection, and
$\mathbf{B} \in \mathbb{R}^{d_{\mathrm{out}} \times r}$ is the
up-projection. Here, $r$ denotes the LoRA rank and is chosen such that $r \ll \min(d_{\mathrm{in}}, d_{\mathrm{out}})$.

\noindent
\textbf{projected LoRA:}
Following the random-projection parameterization introduced in prior work \cite{projected_lora}, we construct the LoRA factors by projecting low-dimensional trainable variables through fixed Gaussian random tensors. Concretely, the projection tensors are sampled once and kept frozen, while only the low-dimensional latent variables are optimized using gradient-based training:
\begin{align}
   \label{eq:projected_lora}
    \mathbf{A}
    &=
    \mathcal{G}_{A}
    \times_3
    \mathbf{m}_{A}, \ 
    \mathbf{B} =
    \mathcal{G}_{B}
    \times_3
    \mathbf{m}_{B},
\end{align}
where
\[
\left[\mathcal{G}_{\cdot} \times_3 \mathbf{m}_{\cdot}\right]_{i,j}
=
\sum_{s=1}^{k}
\left[\mathcal{G}_{\cdot}\right]_{i,j,s}
\left[\mathbf{m}_{\cdot}\right]_s
\]
and $ \mathcal{G}_{A} \in \mathbb{R}^{r \times d_{\mathrm{in}} \times k},$ and $ \mathcal{G}_{B} \in \mathbb{R}^{d_{\mathrm{out}} \times r \times k}$ are frozen Gaussian random tensors whose entries are independently sampled as 
\[
\mathcal{G}\sim\mathcal{N}(\mathbf{0},\Sigma^2),
\qquad
\mathbf{0},\Sigma\in\mathbb{R}^{\operatorname{shape}(\mathcal G)},
\qquad
\Sigma_{ijs}=0.02,
\]
while
$\mathbf{m}_{A},\mathbf{m}_{B}\in\mathbb{R}^{k}$ are trainable intrinsic
vectors. Furthermore, $\times_3$ denotes the mode-3 tensor–vector product, contracting the third dimension of the tensor with the corresponding intrinsic vector.

\subsection{Hierarchy-Conditioned Projected LoRA}

In the original projected formulation Eq. \eqref{eq:projected_lora}, $\mathbf{m}_{A}$ is shared by
all tokens. We instead generate a token-specific intrinsic vector from
the hierarchical row and column headers associated with token $t$:
\[
    \mathbf{m}_{t}
    =
    \mathbf{M}_{\mathbf{c}}
    \odot
    \mathbf{H}^{\mathbf{c}}_t
    +
    \mathbf{M}_{\mathbf{r}}
    \odot
    \mathbf{H}^{\mathbf{r}}_t,
\]
where, for table tokens, $\mathbf{H}^{\mathbf{c}}_t,\mathbf{H}^{\mathbf{r}}_t\in\mathbb{R}^{k}$
denote the encoded column- and row-header hierarchies, respectively, and $\mathbf{M}_{\mathbf{c}},\mathbf{M}_{\mathbf{r}}\in\mathbb{R}^{k}$
are separate trainable vectors for the two hierarchy branches. Here,
$\odot$ denotes the element-wise, or Hadamard product. Each trainable
vector independently rescales the corresponding hierarchy coordinates,
and the two gated representations are added to produce
$\mathbf{m}_t\in\mathbb{R}^{k}$. The effective token-specific down-projection is therefore given by

\[
    \mathbf{A}_t
    =
        \mathcal{G}_{A}
        \times_3
        \left(
        \mathbf{M}_{\mathbf{c}}
        \odot
        \mathbf{H}^{\mathbf{c}}_t
        +
        \mathbf{M}_{\mathbf{r}}
        \odot
        \mathbf{H}^{\mathbf{r}}_t
        \right).
\]

A single frozen tensor $\mathcal{G}_{A}$ is shared by the row and column
branches, while their contributions remain distinct through
$\mathbf{M}_{\mathbf{r}}$ and $\mathbf{M}_{\mathbf{c}}$. 
The up-projection remains shared across tokens:
\[
    \mathbf{B}
    =
    \mathcal{G}_{B}
    \times_3
    \mathbf{m}_{B}.
\]
The resulting adapter update is then given by
\begin{equation}
\label{eq:hierarchical_projected_lora}
    \Delta \mathbf{y}_t
    =
    \frac{\alpha}{r}
    \left(
        \mathcal{G}_{B}
        \times_3
        \mathbf{m}_{B}
    \right)
    \left[
        \mathcal{G}_{A}
        \times_3
        \left(
            \mathbf{M}_{\mathbf{c}}
            \odot
            \mathbf{H}^{\mathbf{c}}_t
            +
            \mathbf{M}_{\mathbf{r}}
            \odot
            \mathbf{H}^{\mathbf{r}}_t
        \right)
    \right]
    \mathbf{x}_t.
\end{equation}

\subsection{Hierarchical Header Representation}
For each table token \(t\), we identify its cell location and the corresponding row and column headers. The column-header hierarchy contains up to \(L_c\) levels, while the row-header hierarchy contains up to \(L_r\) levels. Let \(\mathcal{C}_{t,\ell_c}\) denote the set of tokens in the
column-header cell associated with token \(t\) at column-header level
\(\ell_c\), where
\(\ell_c = 1,\ldots,L_{\mathbf{c}}\). Similarly, let
\(\mathcal{R}_{t,\ell_r}\) denote the set of tokens in the row-header
cell associated with token \(t\) at row-header level \(\ell_r\), where
\(\ell_r = 1,\ldots,L_{\mathbf{r}}\).

To obtain one representation for each header cell at each hierarchy
level, we mean-pool the hidden representations of all tokens within that
header cell:

\[
\begin{aligned}
\bar{\mathbf{h}}^{\mathbf{c}}_{t,\ell_c}
&=
\frac{1}{\left|\mathcal{C}_{t,\ell_c}\right|}
\sum_{s \in \mathcal{C}_{t,\ell_c}}
\mathbf{x}_s,
\qquad
\ell_c = 1,\ldots,L_{\mathbf{c}},
\\
\bar{\mathbf{h}}^{\mathbf{r}}_{t,\ell_r}
&=
\frac{1}{\left|\mathcal{R}_{t,\ell_r}\right|}
\sum_{s \in \mathcal{R}_{t,\ell_r}}
\mathbf{x}_s,
\qquad
\ell_r = 1,\ldots,L_{\mathbf{r}}.
\end{aligned}
\]

For either hierarchy branch, the pooled header-cell representation
satisfies

\[
\bar{\mathbf{h}}^{b}_{t,\ell_b}
\in
\mathbb{R}^{d_{\mathrm{in}}},
\qquad
b \in \{\mathbf{r},\mathbf{c}\}.
\]

The pooled representations from all hierarchy levels are stacked to
form the hierarchy matrix

\[
\bar{\mathbf{H}}^b_t
=
\left[
\bar{\mathbf{h}}^b_{t,1},
\ldots,
\bar{\mathbf{h}}^b_{t,L_b}
\right]
\in
\mathbb{R}^{d_{\mathrm{in}} \times L_b},
\qquad
b \in \{\mathbf{r},\mathbf{c}\},
\]
%
%

We preserve the hierarchy-level dimension and compress only the hidden
dimension. Let $q_b = \frac{k}{L_b}$, where \(k\) is chosen such that it is divisible by both
\(L_{\mathbf{r}}\) and \(L_{\mathbf{c}}\). We apply parameter-free
adaptive average pooling with output size \(q_b \times L_b\):
\[
\mathbf{P}^b_t
=
\operatorname{AdaptiveAvgPool2D}_{(q_b,L_b)}
\left(
\bar{\mathbf{H}}^b_t
\right)
\in
\mathbb{R}^{q_b \times L_b}.
\]

Because the hierarchy dimension remains \(L_b\),
the hierarchy levels stay separate. Only the hidden dimension is reduced from $d_{\mathrm{in}}$ to $q_b$.
The pooled matrix is then vectorized:

\[
\widetilde{\mathbf{H}}^b_t
=
\operatorname{vec}
\left(
\mathbf{P}^b_t
\right)
\in
\mathbb{R}^{q_b L_b}
=
\mathbb{R}^{k}.
\]

Finally, the resulting vector is normalized by its root-mean-square
magnitude:

\[
\mathbf{H}^b_t
=
\frac{
\widetilde{\mathbf{H}}^b_t
}{
\sqrt{
\frac{1}{k}
\sum_{s=1}^{k}
\left(
\widetilde{H}^b_{t,s}
\right)^2
}
+
\epsilon
},
\quad
b \in \{\mathbf{r},\mathbf{c}\} \quad\text{and}\quad\epsilon>0.
\]

\section{Experiment Design}

\subsection{Datasets and Tasks}

\textbf{PK Dataset:}
The dataset comprises scientific tables and their associated textual context, including captions, footnotes, article titles, and abstracts extracted from the web. We used the publicly available PK-specific web crawler described in \cite{ramachandran2023automated} to download 587 full-text articles in XML format. These articles reported PK data for humans, goats, chickens, rats, horses, dogs, cats, and other species. Relevant XML tags were parsed to extract article titles, abstracts, and complete table content, including data, captions, and footnotes. The extracted tables were normalized to preserve their structure and improve consistency. Multi-row headers were identified within the \texttt{<thead>} tag and concatenated column-wise using a caret symbol (\textasciicircum) as a delimiter to retain the original hierarchy. Merged cells were expanded into separate cells, with their content duplicated across the corresponding span to preserve the table layout. This process yielded 950 PK-related tables from 587 articles, as some articles contained multiple tables. The 950 tables were distributed among four annotators for manual labeling. All annotations were subsequently reviewed by a single senior reviewer to resolve inconsistencies and correct annotation errors. The dataset was randomly divided into training, validation, and test sets, with 570 tables (60\%) used for training, 95 tables (10\%) for validation and hyperparameter tuning, and 285 tables (30\%) for testing. 
The task requires data extraction of prompt-relevant PK information from scientific tables and their contextual text. Given a table, its caption and footnotes, and the title and abstract of the associated article, the model must first determine whether the requested information is present. If it is absent, the model returns an empty table. Otherwise, it generates a structured table containing only the relevant extracted values. Because scientific tables often contain multi-row headers, hierarchical column names, merged cells, and contextual labels, the model must associate each extracted value with its corresponding headers at multiple levels. These include column headers, subheaders, row labels, units, and relevant information from captions or footnotes. Preserving these relationships maintains the semantic meaning of the extracted data rather than treating cells independently. For each prompt, the output follows a fixed schema determined by the requested information. The number of columns may vary across prompts, typically ranging from 8 to 12, depending on the type and amount of information requested. These columns represent the extracted PK data and their header-level context, including PK parameters, values, units, species information, dosage, route of administration, number of doses, and dosing interval.

\textbf{WikiTQ:}
WikiTQ is an open-domain table question-answering dataset comprising 22,033 question--answer pairs over 2,108 semi-structured Wikipedia tables \cite{pasupat2015compositional}. Given a natural-language question and its associated table, the task is to generate one or more values that correctly answer the question. The questions require operations such as lookup, comparison, aggregation, and arithmetic reasoning across tables from diverse domains. WikiTQ tables generally use flat, single-level column headers rather than the multi-level row and column hierarchies common in scientific tables. Because the test tables are disjoint from the training tables, the model must also generalize to unseen headers and schemas. We use WikiTQ in its original question-answering formulation as a general-domain comparison baseline and apply the same normalization procedure used for the PK dataset.

\subsection{Experiment Setup}
We evaluated both datasets with Llama~3.2~3B (\texttt{meta-llama/Llama-3.2-3B-Instruct}), Llama~3.1~8B (\texttt{meta-llama/Llama-3.1-8B-Instruct}), and Gemma~3~27B (\texttt{google/gemma-3-27b-it}) using LoRA, TableLoRA, projected LoRA, and our proposed hierarchy-conditioned projected LoRA. All methods used the same data splits, input serialization, prompts, and output format. For TableLoRA, we followed the publicly available implementation of its 2D structural adaptation (\textit{TableLoRA 2D}), where table tokens are assigned learned row and column embeddings based on their table coordinates \cite{he2025tablelora}. Our comparison therefore reflects the structural component provided in the public implementation. Projected LoRA used frozen Gaussian projection tensors and trained only the low-dimensional intrinsic vectors \cite{projected_lora}. Our method used the same projected formulation but generated token-specific intrinsic vectors from the corresponding row- and column-header hierarchies. Unless otherwise stated, adapters were applied to the key and value projection matrices. We used LoRA rank $r=8$, scaling factor $\alpha=16$, and intrinsic dimension $k=1024$. The maximum column- and row-header hierarchy depths were set to $L_{\mathbf{c}}=2$ and $L_{\mathbf{r}}=2$ for PK dataset, and $L_{\mathbf{c}}=1$, $L_{\mathbf{r}}=1$ for WikiTQ. All pretrained model parameters and random projection tensors were frozen. Models were trained with AdamW for 3 epochs with 5 warmup steps, max grad norm 0.5, and weight decay 0. We used bfloat16 precision, gradient checkpointing, and a cosine learning-rate schedule. All experiments were run on a single NVIDIA GH200 Grace Hopper Superchip with 96\,GB of HBM3 GPU memory. The implementation used PyTorch, Hugging Face Transformers, and a custom PEFT-compatible structural adapter framework.

\begin{table*}[t]
\centering
\caption{Performance comparison on the PK and WikiTQ datasets across three backbone models. HARP-LoRA achieves the strongest results for the 8B and 27B backbones while using substantially fewer trainable parameters than LoRA and TableLoRA 2D. Best results within each backbone are shown in \textbf{bold}, and second-best results are \underline{underlined}. P, R, F1, and Acc denote precision, recall, F1-score, and answer accuracy, respectively.}
\label{tab:main_results}
{%
\begin{tabular}{llcccccc}
\toprule
\multirow{2}{*}{\textbf{Model}}
& \multirow{2}{*}{\textbf{Method}}
& \multicolumn{3}{c}{\textbf{PK Dataset}}
& \multicolumn{1}{c}{\textbf{WikiTQ}}
& \multirow{2}{*}{\textbf{\# Train. Params.}} \\
\cmidrule(lr){3-5}
\cmidrule(lr){6-6}
& & \textbf{P (\%)} & \textbf{R (\%)} & \textbf{F1 (\%)}
& \textbf{Acc (\%)} & \\
\midrule

\multirow{4}{*}{Llama3.2 3B}
& LoRA          & \textbf{53.57} & \textbf{58.76} & \textbf{53.57} & 44.17 & 1.84M (0.057\%) \\
& TableLoRA 2D     & 53.13 & 57.90 & 52.87 & 43.80 & 2.59M (0.080\%) \\
& projected LoRA
                 & 52.26 & 57.34 & 52.20 & \textbf{47.32} & 115K (0.004\%) \\
& HARP-LoRA            & \underline{53.26} & \underline{58.09} & \underline{53.08} & \underline{45.23} & 172K (0.005\%) \\

\midrule

\multirow{4}{*}{Llama3.1 8B}
& LoRA          & 57.12 & \underline{59.98} & 56.22 & 46.59 & 2.62M (0.033\%) \\

& TableLoRA 2D     & 53.54 & 57.25 & 53.27 & 42.06 & 3.48M (0.043\%) \\
& projected LoRA
                & \underline{58.27} & 59.24 & \underline{56.78} & \underline{48.97} & 131K (0.002\%) \\
& HARP-LoRA     & \textbf{59.75} & \textbf{60.11} & \textbf{57.64} & \textbf{53.15}  & 197K (0.002\%) \\

\midrule

\multirow{4}{*}{Gemma3 27B}
& LoRA          & 76.32 & 69.02 & 69.05 & \underline{55.73} & 8.36M (0.030\%) \\
& TableLoRA 2D     & 76.55 & \underline{69.14} & 69.19 & 53.67 & 11.82M (0.043\%) \\
& projected LoRA
                & \underline{77.60} & \textbf{69.20} & \underline{69.85} & 55.31 & 365K (0.001\%) \\
& HARP-LoRA          & \textbf{79.11} & 68.80 & \textbf{70.63} & \textbf{56.87} & 547K (0.002\%) \\

\bottomrule
\end{tabular}%
}
\end{table*}

\subsection{Evaluation Metric}
\textbf{PK Dataset:} To evaluate PK-parameter extraction, we used a metric similar to that introduced in \cite{sholehrasa2025autopk}. Each table was annotated with the PK information targeted by the corresponding prompt and its ground-truth output table, including PK values and contextual fields such as dosage, route of administration, species, drug, PK parameter, units, and header-level information. The generated and reference tables were first normalized and cleaned. Generated columns were aligned with reference columns using normalized Levenshtein similarity between column headers, while rows were aligned using a similarity-based assignment procedure over their cell contents. Missing reference rows or columns were represented by empty generated entries, whereas unmatched generated rows and columns were treated as additional predictions. Each non-empty reference cell was then compared with its aligned generated cell. Levenshtein similarity was used for numerical values, PK parameter names, dosage, route of administration, and matrix or commodity fields, while cosine similarity between domain-specific text embeddings was used for fields such as drug and animal names. Following the evaluation settings in AutoPK \cite{sholehrasa2025autopk}, we used a normalized Levenshtein threshold of 0.75 for column alignment, a row-similarity threshold of 0.5, and a cell-level similarity threshold of 0.8. Incorrect or hallucinated generated cells, including cells in extra rows or columns, were counted as false positives, while missing reference cells were counted as false negatives. The true-positive, false-positive, and false-negative counts were pooled across all evaluated tables. Precision, recall, and F1-score were computed for each evaluation run and then averaged separately across three runs on the PK dataset. Because these metrics were averaged independently, the reported mean F1-score is not necessarily equal to the harmonic mean of the reported mean precision and recall.

\textbf{WikiTQ:} We follow the standard evaluation protocol and report answer accuracy \cite{pasupat2015compositional} by comparing each generated answer with its corresponding ground-truth answer. A prediction was counted as correct when the generated answer matched the expected answer.

\section{Results}

\subsection{Main Results}

Table~\ref{tab:main_results} compares HARP-LoRA with LoRA, TableLoRA 2D, and projected LoRA across the PK and WikiTQ datasets. Overall, the benefit of header-conditioned adaptation is more consistent for the evaluated 8B and 27B backbones. For Llama~3.2 3B, HARP-LoRA achieves the second-best performance on both datasets, while outperforming TableLoRA 2D and projected LoRA on PK F1. This indicates that hierarchy-conditioned adaptation provides useful structural information even at the smallest model scale, with particularly clear benefits on the hierarchy-rich PK task.

The improvement becomes clearer with Llama~3.1 8B. HARP-LoRA achieves the best PK F1-score of 57.64\% and the highest WikiTQ accuracy of 53.15\%, improving over projected LoRA by 0.86 and 4.18 points, respectively. These gains are obtained with only 197K trainable parameters, compared with 2.62M for LoRA and 3.48M for TableLoRA 2D. This shows that the added structural conditioning does not require a large increase in trainable capacity. Notably, WikiTQ uses only a single header level, suggesting that header-conditioned adaptation remains useful even when the table does not contain deep hierarchies.

With Gemma~3 27B, HARP-LoRA achieves the strongest overall performance among all evaluated model--method combinations. On the PK dataset, it reaches an F1-score of 70.63\% and a precision of 79.11\%, while on WikiTQ it obtains the highest accuracy of 56.87\%. HARP-LoRA outperforms projected LoRA on both PK F1-score and WikiTQ accuracy while adding only a modest number of trainable parameters. Together with the 8B results, this indicates that the gains are not solely due to the projected low-dimensional parameterization, but also to conditioning the adapter on table header information. The results across the two datasets further suggest that this conditioning is useful in two settings: hierarchy-rich scientific tables, where multi-level headers define the meaning of values, and flatter general-domain tables, where even single-level headers still provide useful structural context.

\subsection{Effect of Intrinsic Dimension}

Table~\ref{tab:intrinsic_dimension} examines the effect of the intrinsic dimension $k$ on HARP-LoRA using Llama~3.1 8B. Increasing $k$ from 256 to 512 produces a large improvement in F1-score, from 30.68 to 55.23. Performance improves further to 57.64 at $k=1024$, which is the best result among the evaluated settings. However, increasing the intrinsic dimension to 2048 slightly reduces the F1-score to 56.79. These results indicate that a larger intrinsic space provides greater capacity to represent the hierarchical row- and column-header information, but the benefit saturates beyond $k=1024$.

\begin{table}[t]
\centering
\caption{Effect of the intrinsic dimension $k$ on the PK data F1-score for HARP-LoRA with Llama~3.1 8B. Performance improves substantially as $k$ increases from 256 to 1024, with the best F1-score achieved at {k}=1024.
}
\label{tab:intrinsic_dimension}
\begin{tabular}{cc}
\toprule
\textbf{Intrinsic Dimension $k$} & \textbf{F1 (\%)} \\
\midrule
  256 & 30.68 \\
  512 & 55.23 \\
 1024 & \textbf{57.64} \\
 2048 & 56.79 \\
\bottomrule
\end{tabular}
\end{table}




\subsection{Ablation Study}

Table~\ref{tab:ablation} evaluates the contribution of each component of HARP-LoRA on the PK dataset using Llama~3.1 8B. The full model achieves the highest F1-score of 57.64, while removing any individual component leads to lower performance. Excluding the row-header hierarchy decreases the F1-score to 56.91, whereas removing the column-header hierarchy results in a larger drop to 56.36. This suggests that both structural sources are useful, with column-header information contributing more strongly in this setting. Removing level-preserving pooling reduces the F1-score to 56.58, indicating that retaining the separation between hierarchy levels is beneficial. Omitting RMS normalization also lowers performance to 56.97, showing that normalization provides a modest but consistent improvement. Finally, replacing the token-specific hierarchy-conditioned intrinsic vector with a shared intrinsic vector, equivalent to projected LoRA, yields an F1-score of 56.78. The 0.86-point improvement of the full model over this variant supports the value of conditioning the adapter on the row- and column-header hierarchies.

\begin{table}[ht]
\centering
\caption{Ablation results on the PK dataset using Llama~3.1 8B. Removing any component reduces the F1-score, with the largest degradation observed when the column-header hierarchy is excluded. The full HARP-LoRA model achieves the best performance.}
\label{tab:ablation}
\small
\setlength{\tabcolsep}{8pt}
\renewcommand{\arraystretch}{1.12}
\begin{tabular}{@{}lc@{}}
\toprule
\textbf{Configuration} & \textbf{F1 Score (\%)} \\
\midrule
Full model
    & \textbf{57.64} \\
Without row-header hierarchy
    & 56.91 \\
Without column-header hierarchy
    &  56.36 \\
Without level-preserving pooling
    &  56.58 \\
Without RMS normalization
    &  56.97 \\
Shared intrinsic vector (projected LoRA)
    &  56.78 \\
\bottomrule
\end{tabular}
\end{table}

\section{Limitations and Future Work}
Compared with projected LoRA, HARP-LoRA has a higher forward-pass cost because its down-projection is token-specific. Let $N$ denote the number of input tokens (sequence length). Constructing the down-projection requires $\mathcal{O}(r d_{\mathrm{in}}k)$ for projected LoRA, since a single projection is shared across all tokens, whereas HARP-LoRA requires $\mathcal{O}(N r d_{\mathrm{in}}k)$. This overhead can become substantial for long sequences or large intrinsic dimensions. Future work will investigate caching and reusing projections for tokens with shared structural contexts.

The current method focuses on reasoning over tables whose cell coordinates and header relationships are already available from preprocessing. Therefore, table detection and structural reconstruction from raw documents are outside the scope of the present framework. In addition, the intrinsic dimension $k$ must be divisible by the selected row- and column-hierarchy depths, which slightly constrains the model configuration. Future work could integrate automatic table-structure recognition and explore more flexible representations that support variable hierarchy depths.

Finally, the current framework is evaluated on text-based tables and does not directly process tables embedded in PDFs or images. Future studies will extend the method to multimodal document settings, automatically infer header hierarchies, and evaluate its generalizability across broader scientific and industrial domains.

\section{Conclusion}

This work introduced HARP-LoRA, a hierarchy-conditioned parameter-efficient fine-tuning method for scientific table understanding. Unlike LoRA and projected LoRA, HARP-LoRA generates token-specific intrinsic vectors from the corresponding row- and column-header hierarchies. This enables the adapter to incorporate the structural context that determines the meaning of table values while keeping the pretrained model and frozen random projection tensors unchanged. Experiments on 950 manually annotated PK tables and WikiTQ show that HARP-LoRA is especially effective for larger language models. It achieves the best overall results with Llama~3.1 8B and Gemma~3 27B while using only 172K--547K trainable parameters, substantially fewer than LoRA and TableLoRA. The results and ablation study further indicate that both header hierarchies, level-preserving pooling, normalization, and token-specific conditioning contribute to performance. Overall, these findings demonstrate that hierarchical table structure can be incorporated directly into lightweight adapters to improve scientific information extraction and general table reasoning. Future work will focus on reducing the computational cost of token-specific down-projections, automatically recovering table structures from raw documents, and extending the framework to multimodal tables in PDFs and images.

\section*{Acknowledgments}

This work was supported by the USDA via the FARAD program (Award No.: 2022-41480-38135, 2023-41480-41034, 2024-41480-43679, and 2025-41480-45282) and its support for the 1DATA Consortium at Kansas State University.

This work was supported by funding from the Kansas Water Institute at Kansas State University.

Research reported in this publication was also partially supported by the Cognitive and Neurobiological Approaches to Plasticity (CNAP) Center of Biomedical Research Excellence (COBRE) of the National Institutes of Health under grant number P20GM113109. The content is solely the responsibility of the authors and does not necessarily represent the official views of the National Institutes of Health.

This work used the Delta and DeltaAI resources at the National Center for Supercomputing Applications (NCSA) through ACCESS allocation CIS260493. ACCESS is supported by the U.S. National Science Foundation under Grants No. 2138259, 2138286, 2138307, 2137603, and 2138296.

\bibliographystyle{IEEEtran}
\bibliography{references}

@inproceedings{balazy2405lora,
  title={{Lora-XS}: Low-rank adaptation with extremely small number of parameters},
  author={Ba{\l}azy, Klaudia and Banaei, Mohammadreza and Aberer, Karl and Tabor, Jacek},
  booktitle={ECAI 2025: 28th European Conference on Artificial Intelligence 25-30 October 2025, Bologna, Italy-Including 14th Conference on Prestigious Applications of Intelligent Systems (PAIS 2025) Proceedings},
  year={2025},
}

@inproceedings{he2025tablelora,
  title={{TableLoRA}: Low-rank adaptation on table structure understanding for large language models},
  author={He, Xinyi and Liu, Yihao and Zhou, Mengyu and He, Yeye and Dong, Haoyu and Han, Shi and Yuan, Zejian and Zhang, Dongmei},
  booktitle={Proc. of the 63rd Annual Meeting of the Association for Comp. Linguistics (Volume 1: Long Papers)},
  year={2025}
}

@inproceedings{sholehrasa2025autopk,
  title={Autopk: Leveraging llms and a hybrid similarity metric for advanced retrieval of pharmacokinetic data from complex tables and documents},
  author={Sholehrasa, Hossein and Ghanaatian, Amirhossein and Caragea, Doina and Tell, Lisa A and Riviere, Jim E and Jaberi-Douraki, Majid},
  booktitle={2025 IEEE 37th International Conference on Tools with Artificial Intelligence (ICTAI)},
  pages={338--346},
  year={2025},
  organization={IEEE}
}

@inproceedings{yin2020tabert,
  title={TaBERT: Pretraining for joint understanding of textual and tabular data},
  author={Yin, Pengcheng and Neubig, Graham and Yih, Wen-tau and Riedel, Sebastian},
  booktitle={Proc. of the 58th Annual Meeting of the Association for Comp. Linguistics},
  pages={8413--8426},
  year={2020}
}

@inproceedings{herzig2020tapas,
  title={TaPas: Weakly supervised table parsing via pre-training},
  author={Herzig, Jonathan and Nowak, Pawel Krzysztof and M{\"u}ller, Thomas and Piccinno, Francesco and Eisenschlos, Julian},
  booktitle={Proc. of the 58th annual meeting of the ACL},
  pages={4320--4333},
  year={2020}
}

@article{deng2022turl,
  title={Turl: Table understanding through representation learning},
  author={Deng, Xiang and Sun, Huan and Lees, Alyssa and Wu, You and Yu, Cong},
  journal={ACM SIGMOD Record},
  year={2022},
  publisher={ACM New York, NY, USA}
}

@inproceedings{jiang2022omnitab,
  title={OmniTab: Pretraining with natural and synthetic data for few-shot table-based question answering},
  author={Jiang, Zhengbao and Mao, Yi and He, Pengcheng and Neubig, Graham and Chen, Weizhu},
  booktitle={Proceedings of the 2022 Conference of the North American Chapter of the Association for Computational Linguistics: Human Language Technologies},
  pages={932--942},
  year={2022}
}

@inproceedings{zhang2024tablellama,
  title={Tablellama: Towards open large generalist models for tables},
  author={Zhang, Tianshu and Yue, Xiang and Li, Yifei and Sun, Huan},
  booktitle={Proc. of the 2024 conf. of the North American chapter of the ACL: human language technologies (volume 1: long papers)},
  pages={6024--6044},
  year={2024}
}

@inproceedings{
   zhang2023adalora,
   title={Adaptive Budget Allocation for Parameter-Efficient Fine-Tuning },
   author={Qingru Zhang and Minshuo Chen and Alexander Bukharin and Pengcheng He and Yu Cheng and Weizhu Chen and Tuo Zhao},
   booktitle={The Eleventh International Conference on Learning Representations },
   year={2023},
}

@Misc{peft,
  title =        {{PEFT}: State-of-the-art Parameter-Efficient Fine-Tuning methods},
  author =       {Sourab Mangrulkar and Sylvain Gugger and Lysandre Debut and Younes Belkada and Sayak Paul and Benjamin Bossan and Marian Tietz},
  howpublished = {\url{https://github.com/huggingface/peft}},
  year =         {2022}
}

@article{hu2022lora,
  title={{LoRA}: Low-rank adaptation of large language models.},
  author={Hu, Edward J and Shen, Yelong and Wallis, Phillip and Allen-Zhu, Zeyuan and Li, Yuanzhi and Wang, Shean and Wang, Liang and Chen, Weizhu and others},
  journal={International Conference on Learning Representations},
  year={2022}
}

@inproceedings{cheng2022hitab,
  title={Hitab: A hierarchical table dataset for question answering and natural language generation},
  author={Cheng, Zhoujun and Dong, Haoyu and Wang, Zhiruo and Jia, Ran and Guo, Jiaqi and Gao, Yan and Han, Shi and Lou, Jian-Guang and Zhang, Dongmei},
  booktitle={Proceedings of the 60th Annual Meeting of the Association for Computational Linguistics},
  year={2022}
}

@book{riviere2018veterinary,
  title={Veterinary pharmacology and therapeutics},
  author={Riviere, Jim E and Papich, Mark G},
  year={2018},
  publisher={John Wiley \& Sons}
}

@inproceedings{maunsell2021farad,
  title={FARAD: How we respond to withdrawal inquiries},
  author={Maunsell, Fiona and Baynes, Ron and Davis, Jennifer and Foster, Derek and Jaberi-Douraki, Majid and Riviere, Jim and Tell, Lisa},
  booktitle={American Association of Bovine Practitioners Conference Proceedings},
  pages={9--11},
  year={2021}
}

@article{ramachandran2023automated,
  title={An automated customizable live web crawler for curation of comparative pharmacokinetic data: an intelligent compilation of research-based comprehensive article repository},
  author={Ramachandran, Remya Ampadi and Tell, Lisa A and Rai, Sidharth and Gedara, Nuwan Indika Millagaha and Xu, Xuan and Riviere, Jim E and Jaberi-Douraki, Majid},
  journal={Pharmaceutics},
  volume={15},
  number={5},
  pages={1384},
  year={2023}
}

@article{smith2025automated,
  title={An automated classification pipeline for tables in pharmacokinetic literature},
  author={Smith, Victoria C and Gonzalez Hernandez, Ferran and Wattanakul, Thanaporn and Chotsiri, Palang and Cordero, Jos{\'e} Antonio and Ballester, Maria Rosa and Duran, M{\`a}rius and Fanlo Escudero, Olga and Lilaonitkul, Watjana and Standing, Joseph F and others},
  journal={Scientific Reports},
  year={2025},
  publisher={Nature Publishing Group UK London}
}

@article{golmohammadi2025comprehensive,
  title={Comprehensive assessment of adverse event profiles associated with bispecific antibodies in multiple myeloma},
  author={Golmohammadi, Mobina and Raza, Shahzad and Albayyadhi, Maram and Sholehrasa, Hossein and Khouri, Jack and Williams, Louis and Hansen, Doris K and Moradi, Azam and Xu, Xuan and Albliwi, Moath and others},
  journal={Blood Cancer Journal},
  year={2025},
  publisher={Nature Publishing Group UK London}
}

@article{projected_lora,
  title={Derivative-free optimization for low-rank adaptation in large language models},
  author={Jin, Feihu and Liu, Yifan and Tan, Ying},
  journal={IEEE/ACM Transactions on Audio, Speech, and Language Processing},
  volume={32},
  pages={4607--4616},
  year={2024},
  publisher={IEEE}
}

@inproceedings{pasupat2015compositional,
    title = "Compositional Semantic Parsing on Semi-Structured Tables",
    author = "Pasupat, Panupong  and
      Liang, Percy",
    editor = "Zong, Chengqing  and
      Strube, Michael",
    booktitle = "Proceedings of the 53rd Annual Meeting of the ACL and the 7th Int. Joint Conf. on NLP (Volume 1: Long Papers)",
    month = jul,
    year = "2015",
    address = "Beijing, China",
    doi = "10.3115/v1/P15-1142",
    pages = "1470--1480"
}

@article{li2024tablegpt,
  title={Table-gpt: Table fine-tuned gpt for diverse table tasks},
  author={Li, Peng and He, Yeye and Yashar, Dror and Cui, Weiwei and Ge, Song and Zhang, Haidong and Rifinski Fainman, Danielle and Zhang, Dongmei and Chaudhuri, Surajit},
  journal={Proceedings of the ACM on Management of Data},
  volume={2},
  number={3},
  pages={1--28},
  year={2024},
  publisher={ACM New York, NY, USA}
}

@inproceedings{zhang2025tablellm,
  title={Tablellm: Enabling tabular data manipulation by llms in real office usage scenarios},
  author={Zhang, Xiaokang and Luo, Sijia and Zhang, Bohan and Ma, Zeyao and Zhang, Jing and Li, Yang and Li, Guanlin and Yao, Zijun and Xu, Kangli and Zhou, Jinchang and others},
  booktitle={Findings of the Association for Computational Linguistics: ACL 2025},
  pages={10315--10344},
  year={2025}
}

@inproceedings{chen2019tabfact,
  title={{TabFact}: A large-scale dataset for table-based fact verification},
  author={Chen, Wenhu and Wang, Hongmin and Chen, Jianshu and Zhang, Yunkai and Wang, Hong and Li, Shiyang and Zhou, Xiyou and Wang, William Yang},
  booktitle={International Conference on Learning Representations},
  year={2020}
}

@inproceedings{sui2024tablemeetsllm,
author = {Sui, Yuan and Zhou, Mengyu and Zhou, Mingjie and Han, Shi and Zhang, Dongmei},
title = {Table Meets LLM: Can Large Language Models Understand Structured Table Data? A Benchmark and Empirical Study},
year = {2024},
isbn = {9798400703713},
publisher = {Association for Computing Machinery},
address = {New York, NY, USA},
url = {https://doi.org/10.1145/3616855.3635752},
doi = {10.1145/3616855.3635752},
booktitle = {Proceedings of the 17th ACM International Conference on Web Search and Data Mining},
pages = {645–654},
numpages = {10},
location = {Merida, Mexico},
series = {WSDM '24}
}

@misc{farad,
  author = {{Food Animal Residue Avoidance Databank (FARAD)}},
  title = {FARAD - Food Animal Residue Avoidance Databank},
  howpublished = {\url{http://www.farad.org/}},
  note = {Accessed: 2026-09-25}
}

@inproceedings{wang2021tuta,
  title={Tuta: Tree-based transformers for generally structured table pre-training},
  author={Wang, Zhiruo and Dong, Haoyu and Jia, Ran and Li, Jia and Fu, Zhiyi and Han, Shi and Zhang, Dongmei},
  booktitle={Proceedings of the 27th ACM SIGKDD Conference on Knowledge Discovery \& Data Mining},
  pages={1780--1790},
  year={2021}
}

@article{meng2024pissa,
  title={Pissa: Principal singular values and singular vectors adaptation of large language models},
  author={Meng, Fanxu and Wang, Zhaohui and Zhang, Muhan},
  journal={Advances in Neural Information Processing Systems},
  year={2024}
}

@article{pkdb,
  title={PK-DB: pharmacokinetics database for individualized and stratified computational modeling},
  author={Grzegorzewski, Jan and Brandhorst, Janosch and Green, Kathleen and Eleftheriadou, Dimitra and Duport, Yannick and Barthorscht, Florian and K{\"o}ller, Adrian and Ke, Danny Yu Jia and De Angelis, Sara and K{\"o}nig, Matthias},
  journal={Nucleic acids research},
  volume={49},
  number={D1},
  pages={D1358--D1364},
  year={2021},
  publisher={Oxford University Press}
}

@inproceedings{sholehrasa2025predictive,
  title={Predictive modeling and explainable ai for veterinary safety profiles, residue assessment, and health outcomes using real-world data and physicochemical properties},
  author={Sholehrasa, Hossein and Xu, Xuan and Caragea, Doina and Riviere, Jim E and Jaberi-Douraki, Majid},
  booktitle={2025 IEEE International Conference on Big Data (BigData)},
  pages={2043--2052},
  year={2025},
  organization={IEEE}
}

@article{jaberi2025hysim,
  title={Hysim-llm: Embedding-weighted fine-tuning bounds and manifold denoising for domain-adapted llms},
  author={Jaberi-Douraki, Majid and Sholehrasa, Hossein and Xu, Xuan and Ramachandran, Remya Ampadi},
  journal={arXiv preprint arXiv:2510.07796},
  year={2025}
}

\end{document}